\documentclass[wcp]{jmlr}
\usepackage{longtable}
\usepackage[british]{babel}
\usepackage[autostyle]{csquotes}
\usepackage{caption}
\usepackage{booktabs}
\makeatletter
\let\Ginclude@graphics\@org@Ginclude@graphics 
\makeatother
\title[Towards Benchmarking XAI-methods]{Towards Benchmarking Explainable Artificial Intelligence Methods}
  \author{\Name{Lars Holmberg} \Email{lars.holmberg@mau.se}\\
  \addr Department of Computer Science and Media Technology, Malmö University, Sweden
}
\begin{document}
\maketitle
\begin{abstract}
The currently dominating artificial intelligence and machine learning technology, neural networks, builds on inductive statistical learning. Neural networks of today are information processing systems void of understanding and reasoning capabilities, consequently, they cannot explain promoted decisions in a humanly valid form. In this work, we revisit and use fundamental philosophy of science theories as an analytical lens with the goal of revealing, what can be expected, and more importantly, not expected, from methods that aim to explain decisions promoted by a neural network. By conducting a case study we investigate a selection of explainability method's performance over two mundane domains, animals and headgear. Through our study, we lay bare that the usefulness of these methods relies on human domain knowledge and our ability to understand, generalise and reason. The explainability methods can be useful when the goal is to gain further insights into a trained neural network's strengths and weaknesses. If our aim instead is to use these explainability methods to promote actionable decisions or build trust in ML-models they need to be less ambiguous than they are today. In this work, we conclude from our study, that benchmarking explainability methods, is a central quest towards trustworthy artificial intelligence and machine learning.  
\end{abstract}


\section{Introduction}
Digitalisation influences all parts of society. In its heart resides Artificial Intelligence (AI) and Machine Learning (ML), technologies with roots in natural science and a third-person objectivising stance~\cite{Grimm2016}. Research in the area then inherits values that are concerned with ground truth, optimising class probability, minimising bias in training data and mitigating consequences of data drift. This approach, useful when the target domain is static and well-defined, is not well suited to promote decisions\footnote{The output from an ML system in the form of classification, recommendation, prediction, proposed decisions or action} in a real-world context~\cite{Chollet2019}. These ML/AI systems are information processing systems that due to their complexity becomes black boxes~\cite{Lipton2016} that promote decisions without presenting reasons in a human-understandable form. Answering a \textit{how-question}, by associating inputs to an ML-system with a promoted decision, compared to, answering a \textit{why-question}, actionable in the real world, are two very different challenges. In later years negative consequences related to this information processing approach, void of reasoning and understanding, has become increasingly apparent~\cite{Hutchinson2019,couldry}.

The gap between reality and the fragment of reality modelled by an ML/AI System can be expressed in terms of explanandum overlap. There is then a significant explanandum overlap in medical image analyse, where the problem is hermeneutic and concerned with interpreting, analysing and naming concepts in images produced by an instrument that visualises concepts unreachable for human senses. Areas with limited explanandum overlap, on the other hand, are, for example, image search on the internet for abstract concepts like freedom, sexual orientation, gender or ethnicity. The explanandum overlap is then limited with respect to: how these socially constructed concepts are perceived by humans, and how and in what way they are visually represented in an image search. Explanations based on an image search can then only communicate very limited visual queues to these concepts. Datasets, used for training and benchmarking, are, in addition to this, contextually and culturally situated and consequently come with inherited and embedded value systems. These enormous and inconceivable datasets then risk being normative and part of conserving and obscuring values if they are used downstream in ML-models deployed in a real-world setting~\cite{feifei2020,gebru}.

\begin{figure}[!t]
\begin{center}
\includegraphics[width=1.0\textwidth]{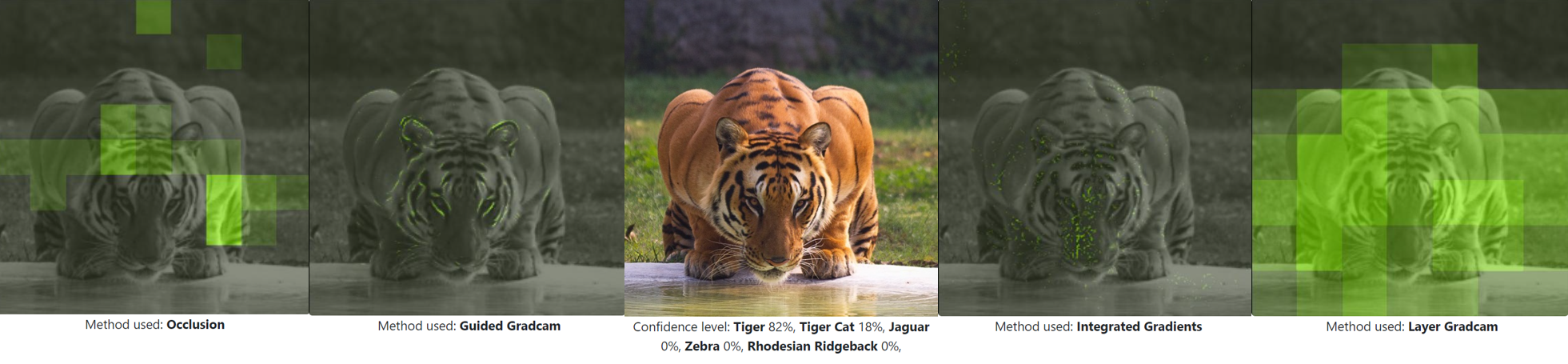}
\caption{Why is this image classified as a \textquote*{Tiger}? \\ \tiny\textsuperscript{*} Photo by Ratanjot Singh on Unsplash}
  \label{fig:teaser}
\end{center}
\end{figure}

The work presented here then focus on this gap, between the fragment of reality modelled by the ML/AI system and the real world in which a promoted decision can be useful and actionable. We adhere to a Human In Command approach (HIC)~\cite{holmberglic} with the goal of positioning the Machine In The Loop (MITL) and thus presenting an alternative to a technology trend that position Humans In The Loop (HITL) as a resource that risks being treated instrumentally~\cite{couldry}. Moving the agency in this direction opens up for a type of ML/AI where humans have agency and can value certain ends and goals as good and prosperous for humanity~\cite{Vallor2016}.

This paper focus on inductive statistical learning approaches used in the currently dominating ML technology, neural networks. This is a technology that can learn from raw input data in the form of images, sound or text~\cite{Lecun2015}. We selected this ML/AI approach for our analyse since knowledge priors embedded in the system originate primarily from the selection of training data and labels which, make our analyse simpler. We focus on classification tasks for images picturing mundane visual objects as input data and we select those images so the explanandum available to the human and the ML-system overlap. We thereby construct a setting in which a human with domain knowledge (henceforth domain experts) can assess reasons for a decision presented by the ML system. 

Our focus is to research if available eXplainable Artificial Intelligence (XAI) methods are useful when the goal is to understand a trained ML-model's decisions. By selecting a mundane domain we build on a tradition of benchmarks, but, instead of focusing on average-case performance for a static dataset (ImageNet1000)~\cite{Deng2010}, our focus is on the usefulness of XAI-methods given a pretrained ML-model (ResNet)~\cite{He2016} and carefully selected images. We, therefore, formulate the following research question:
\begin{itemize}
    \item{Are local XAI-methods capable of communicating reasons for a proposed decision in a human understandable form?}
\end{itemize}

To address this we study five local XAI-methods that utilise saliency maps to accentuate parts of the image as important for a singular promoted decision- [See Figure~\ref{fig:teaser}]). These explanation methods cannot give a comprehensive summary of all reasons for a decision, instead, they aim to present evidence from the part of the explanandum reachable for the specific XAI-method in question in a human-understandable form.

In the study, we find that the XAI-methods delivers vague and imprecise evidence and this leaves interpretation and judging the evidence presented to human abilities. Notwithstanding this, we believe that the tested XAI-methods, especially when used in an ensemble, can deliver valuable insights on strengths and weaknesses related to the ML-model's capabilities. We also find that the XAI-methods has to be used strategically since they also have their different strengths and weaknesses and therefore deliver complementing insights in relation to the ML-model's Internal Knowledge Representation (IKR).         

In the background section that follows, we present theories related to scientific explanations, inductive statistical learning and concepts. This is followed by a section on the methodological approach leading to our study results. The article ends with a discussion section that contextualises our conclusions. We then end the article by concluding the results.

\section{Background}
Human understanding of natural phenomena builds on the uniformity principle which implies that instances of which we have no experience must resemble those we had experience from~\cite{Hume1896}. The uniformity principle cannot be justified from a scientific perspective and we, therefore, need to define non-inductive reasons to classify an explanation as scientific. To be scientifically valid we need a theory that allows for deductive reasoning that consequently can be used to predict future events without relying on induction. This is then important for machine learning, based on induction, since without the ability to falsify the system and scientifically explain decisions there is no demarcation between these systems and pseudoscience. 

In this work we use a form of scientific explanations, Deductive-Nomological (DN)~\cite{hempel}, as our analytical lens to better understand what can be expected, and not expected, from XAI-methods used in conjunction with neural networks. DN-explanations are used throughout science as a blueprint for a scientific explanations and can then serve us when we discuss what we can expect, epistemologically, from explanations that reflect the inductive statistical learning process used when neural networks are trained.  

A full DN-explanation uses a theory to create a model of the phenomena in question, this model then describes kinds and entities that populate the model~\cite{Overton} (We will in the following work denote this type of model, DN-model, to distinguish it from an ML-model). Measurements in the form of data are then, if the DN-model describes the phenomena well, aligned to predictions produced by the DN-model. A scientific explanation can answer a why-question, DN nomenclature for this is that a why-question establish an explanandum from which explanans, or evidence for a decision, can be selected to form a scientific explanation that answers the why-question. At least one of these evidences needs to be a law deduced from the DN-model~\cite{woodward2019}.  

In algorithmic ML, a theory is used to select an algorithm that can encompass the phenomena in question and consequently the DN-model and the ML-model overlaps. A neural network is, instead of using a preselected algorithm, limited by the universal approximation theorem~\cite{Hornik1991} which imply that the resulting ML-model used for prediction is formed during training based on the training data and labels. Consequently, there exists no model in a DN sense, no model that builds on a theory that can be falsified. From this follows that the neural network is incapable of producing a scientifically valid explanation since they build on a theory and deductive reasoning.

In this work, we denote all human knowledge added to the ML-system as knowledge priors. This can be choice of domain, algorithm selection, data augmentation, feature engineering, training data selection, etc. For a neural network knowledge priors are mainly added based on domain selection and input data selection. Neural networks then build internal knowledge representations from raw data (images, sound or text) and labels, the function used for prediction is then created in an inductive statistical manner by exposing the neural network to a large amount of training data~\cite{Lecun2015}. 

The often assumed prerequisite for neural network training is that data and labels are independent and identically distributed (i.i.d), a presumption that is very challenging to fulfil in a real-world setting~\cite{scholkopf2019}. Data not part of the intended target domain is in our work denoted out of distribution data (o.o.d.). A consequence of the inductive learning process, and the nonexisting DN-model, is that o.o.d. data cannot be identified as external by a neural network. This since the network, cannot, without human help, identify the domain borders. It is then, in a classification problem, not possible to identify new classes, instead, class probability for a promoted decision reflect some aspect of similarity with existing classes.   

Concepts are central to human reasoning and essential for our ability to generalise. We use and need them to explain and make predictions about new objects and situations~\cite{sep-concepts,Murphy2018}. In this work, we aim to use XAI-methods and map the IKRs (Internal Knowledge Representations) in a trained neural network to Human Understandable Concepts (HUC). This to some extent maps to prototype theories and exemplar theories of concepts~\cite{Murphy2016}. Prototype theories are a summary description of concepts (our HUC) and exemplar theories are remembered instance categories (our IKR). Humans’ beliefs, related to concepts and their properties, can be both false and incomplete and, additionally, contain both causal and descriptive factors~\cite{Genone2012}. Concepts are then named using a referent, for example, \textquote*{pferd} or \textquote*{horse} (double quotation marks), these are two referents that points to the same concept, \textquote{horse} (single quotation marks).

An approach used to search for a precise definition of concepts is to specify them as necessary and/or sufficient~\cite{brennan2017}. This approach is used in machine learning both in experimental research~\cite{Wang2020} and to underpin representation learning~\cite{Wang2021}. To exemplify, the concept of \textquote{elephant} can be described using necessary sub-concepts shared with many animals, for example, \textquote{four legs}, \textquote{eyes} and sufficient sub-concepts, for example, \textquote{elephant tusk}. Sufficient and necessary sub-concepts are those that on their own can be used in classification, for example, \textquote{elephant trunk} (\textquote{elephant tusks} are sufficient for classification but not necessary since not all female Indian elephants have tusks). Spurious correlations are relationships between the proposed decision that are prone to change when a system is deployed in a real-world context~\cite{gulrajani2020}. For example, if all elephants are pictured close to \textquote{watering holes} this concept can wrongly be seen by the ML-model as a necessary concept. In one part of our study, we investigate the usefulness of spurious correlations and sufficient and necessary concepts since human insights here can help to identify limitations in the training data in relation to the deployment domain. In this work we use the notion of referents to concepts synonymous with labels, promoted decisions and classifications. 

In recent years there has been a surge in XAI-methods~\cite{Gilpin2019,samek2021} but there are few user studies evaluating these methods in a real-world context~\cite{Tjoa2019}. In the last years, implementations of many of the XAI-methods in accessible packages have been produced, which simplifies user studies. One advantage is that the XAI-methods are implemented by a group of developers and therefore are, from an implementation and code quality perspective, comparable. In this work, we have selected Captum (PyTorch) as a competitive implementation~\cite{captum}. It has then become easier to use XAI-methods, but still, it is not clear if the XAI-methods are mostly developed to give insights into the trained ML-models or if they also are useful for our target users: domain experts. 

For our study we have selected both XAI-methods that are ML-model independent, like Occlusion~\cite{occlusion2014} and those that focus on IKR in layers closer to the promoted decision, two Gradient-weighted Class Activation Mapping (GradCAM)~\cite{Selvaraju2016} methods and a SHapley Additive exPlanations (SHAP)~\cite{Lundberg} method, as well as one method, Integrated Gradients (IG)~\cite{integratedgradient2017}, that weighs activations over all layers in a neural network. XAI-methods, in their original papers, are often presented with best-case performance examples and their usefulness can even be questioned~\cite{Adebayo2018}. We are then interested in evaluating them using pictures from mundane domains and consequently get an idea of their more general usefulness. The underlying reasoning is that the XAI-methods should perform well in a domain that pictures well-known objects if they are to be trusted in more demanding domains.

\begin{figure*}[ht]
\centering
\includegraphics[width=\linewidth]{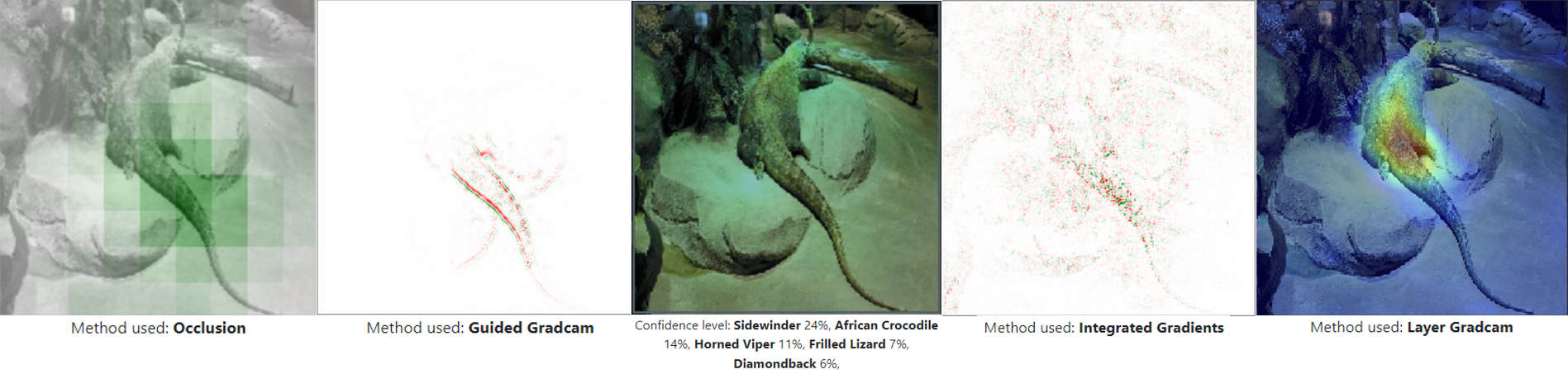}
\caption{Examples of the diverse visual languages used in XAI-methods. The original image and predictions are placed in the middle, flanked by different XAI-methods.}
\label{fig:comparing_methods}
\end{figure*}

\section{Methodology}
We set up a small but targeted case study with the goal of exploring some alternative approaches that ML/AI systems can use to communicate reasons for decisions with humans~\cite{casestudy}. One of the two studies investigated if the notion of sufficient concepts, necessary concepts and spurious correlations can be useful as an analytical tool when the goal is to categorise areas in images accentuated by the XAI-methods. In the other study, we investigated if we can rely on a human intuitive understanding of the XAI-methods explainability capabilities. In both cases, we selected images picturing non-abstract concepts clearly visible in the images (animals and headgear). The studies mixed harder to identify objects with easier and a few images that pictured more than one object. We also selected images picturing objects predicted with class probability around 50\% that. The study participants were not told which classes, within the domain, the system could recognise, they only got information related to the domain they could expect, animals and headgear. By this approach, we aimed towards a baseline study where the results of the two studies give basic indications on the usefulness of the XAI-methods in a mundane domain.  

Since our goal is benchmarking XAI-methods we used well-known datasets and pre-trained models, ImageNet1000 together with a pre-trained ResNet50 model ResNet50, a model that has a reasonably low error rate (less than 5\% in the ILSVRC~\cite{Deng2010} challenge). With this setup, together with a coherent XAI-API, we aimed towards a setting with a clear focus on the usability of the XAI-methods. Around this ML architecture, we created a website containing the two studies. Each study consisted of web pages the participant could navigate back and forth between. On each page one image pictured an object and a prediction together with a possibility to choose an XAI-method to investigate. A form for answers was placed under the images so the participants could select the methods they found most useful, this was accompanied by free text explanation fields. After the study, the participants were asked to summarise their understanding of the ResNet50-models overall capabilities related to the domain in question.

The XAI-methods selected, in their original implementations, use a diverse and to some extent incompatible graphical language to accentuate areas important for a decision in the image. For example, GradCAM images are often visually appealing compared to gradient-based methods (See Figure~\ref{fig:comparing_methods}). Differences lay partly in colour schemes used and if the visual explanations are overlaid on the original image. GradCAM uses, in the original implementation, colour gradients from blue to red to indicate areas that increasingly influence the prediction. Other methods use red colour to indicate areas that influence the attribution negatively~\cite{Wang2020}. In the XAI-method Integrated Gradient, internal gradients for a predicted class are projected on the input image and coloured brownish. These diverse visual explanation languages are then not comparable and each of them needs to be explained to be understood. We discussed, in our research team, what a negative attribution implies and came to the conclusion that it is not intuitive and its usefulness can be questioned. A negative attribution will also be differently interpreted if the promoted decision is perceived as erroneous. For example, what do the concept \textquote{non-cowboy hat} imply for a erroneous prediction of a \textquote*{top hat} as a \textquote*{cowboy hat}. So, instead of adding to the complexity more than needed, we decided to focus on identifying sub-concepts that can be associated with the promoted decision, whether it is perceived as correct or incorrect by the domain expert.   

\begin{figure*}[ht]
\centering
\includegraphics[width=\linewidth]{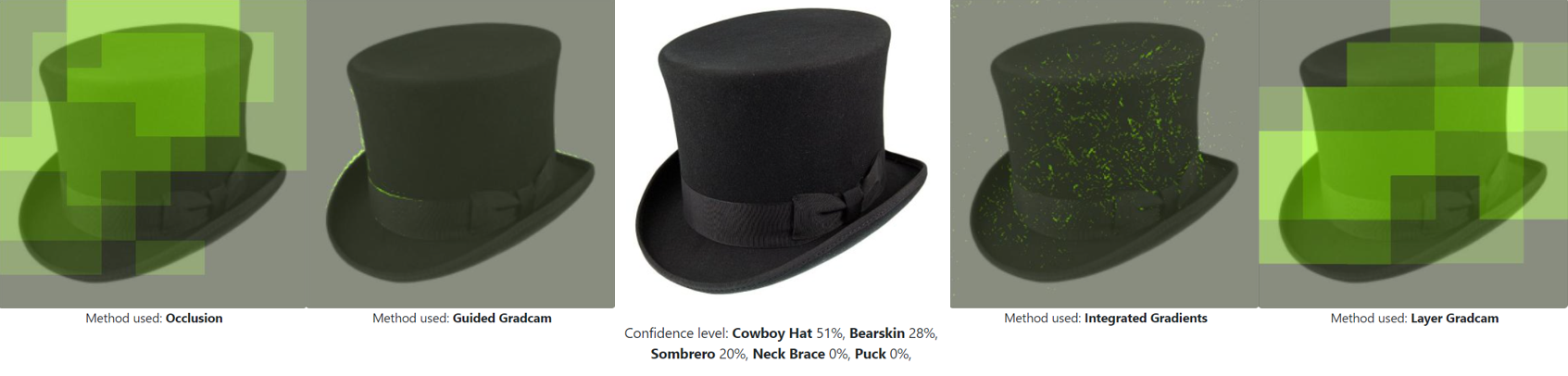}
\caption{Part of the headgear study, visualising areas in images accentuated as important for a promoted decision. The original image and predictions are placed in the middle, flanked by different XAI-methods.}
\label{fig:oversight2}
\end{figure*}

Based on the reasoning above we decided to use a coherent graphical language for the XAI-methods and thereby increase comparability between the methods in a baseline test. We decided on a design with two images, the original image with predictions and another image with the selected XAI-method overlaid on the original image in black and white and slightly opaque. We then used a bright green colour that contrasts to the greyish background and we let opacity of the green colour indicate how important a part of an image is for the promoted decision (See Figure~\ref{fig:teaser} and Figure~\ref{fig:oversight2}).

By keeping many parameters constant we aimed towards a controlled experiment~\cite{Ko2013a} and semi-structured interviews~\cite{Myers2007}. Interviews and text were for the animal study broken down into codes related to sufficient and necessary concepts and mentioning of spurious correlations. In the study, four students from an Ethics and IT bachelor level course participated together with four bachelor exam-students, with a focus on XAI, and two researchers with a background in ML/AI. The age span was 20-40 years and 40\% of the participants were women. Except collecting the form data from the website we interviewed six of the participants in half-hour sessions. From the forms, we could collect some quantitative data on relative measurement of the XAI-methods perceived usefulness.

\section{Result}
One of our two studies investigated whether it is useful, for a human, to categorise visible sub-concepts in images picturing animals as necessary, sufficient and, additionally, the usefulness of the notion of spurious correlations concerning concepts not deemed as necessary or sufficient. Concepts here were then sub-concepts to animal classifications, as, for example, \textquote{watering hole}, \textquote{beak} or \textquote{feather}. For the majority of the users, this approach was not seen as a useful path to evaluate the XAI-methods. Only one participant was cautiously positive, but generally, the usage of this categorisation of concepts was seen as puzzling or even uncomfortable since they are subjective and open for discussion. 

The methods Layer GradCAM and Guided GradCAM were mentioned as the ones that resemble a human approach to the same task closest and thus were deemed as most useful. For example, a somewhat non-sharp image picturing a type of lizard, \textquote{komodo dragon}, was by the ML-model erroneously classified as a type of snake with 24\% class probability (See central image in Figure~\ref{fig:comparing_methods}). In this case, the XAI-methods were useful since they drew attention to the form of the tail as a possible reason. Another example: since~\textquote{horse} is not one of the classes in ImageNet1000 it is not possible to classify for the ML-model. This was disclosed by two users that used the XAI-methods to point towards reasons why the \textquote{horse} was erroneously classified as the dog bread \textquote*{great dane} and thus drew the conclusion that \textquote{horse} must be an o.o.d class. 

The study participants judged that the ML-model predicted correctly for 60\% of the images. In total 64 images were assessed by the participants. It was only in one case a participant used the possibility to refrain from selecting any method. The perceived usefulness for the XAI-methods for the animal study is presented in Table~\ref{tab:usefulness}. The fact that the study participants didn't know which classes the system was trained on led to some confusion. For example, classifying a pond with three ducks as \textquote*{drake} is from a human perspective a questionable focus. 
The use of necessary concepts, sufficient concepts and spurious correlations can probably be useful but our results indicate that the study participants need a theoretical base to make use of these notions. A peculiarity, when the methods are compared, is that when classifications are judged as erroneous Guided GradCAM is perceived as more useful (42\%) than Layer GradCAM (27\%). The relation is switched for decisions judged as correct, where Layer GradCAM (45\%) is perceived as more useful than Guided GradCAM (31\%). 

\begin{table}[t]
\centering
  \caption{The participant's subjective and relative assessment of the usefulness of XAI-methods. TP (True Positives) denotes usefulness of XAI methods for predictions perceived as correct. FP (False Positives) denotes the usefulness of methods for predictions perceived as erroneous (The participant could choose multiple XAI methods for each proposed prediction).}
  \label{tab:usefulness}
  \begin{tabular}{l|lll|lll}
    \toprule
    \multicolumn{1}{l|}{}&\multicolumn{3}{c|}{Animal study} & \multicolumn{3}{c}{Headgear study}\\
    \midrule
    XAI-method&All&TP&FP&All&TP&FP\\
    \midrule
    Occlusion & 19\% & 18\% & 19\% & 29\%& 29\% & 31\%  \\
    Guided GradCAM & 35\% & 31\% & 42\% & 27\% & 27\% & 27\%\\
    Integrated Gradients & 7\% & 5\% & 12\% & 5\% & 5\% & 6\% \\
    Layer GradCAM & 38\% & 45\% & 27\% & 36\% & 38\% & 32\% \\
    Gradient SHAP & n/a & n/a & n/a & 3\% & 2\% & 4\%  \\
  \bottomrule
\end{tabular}
\end{table}

\label{sec:study on headgear}
The other part of the study focused on headgear and the six classes in ImageNet1000: \textquote{sombrero}, \textquote{cowboy hat}, \textquote{bathing cap}, \textquote{crash helmet}, \textquote{bonnet}, \textquote{shower cap}. We used the same XAI-methods as in the animal study but added Gradient SHAP, the reason being that SHAP methods are referenced in many XAI-papers and, based on experiences from the animal study, adding one more method would not be confusing.

The study participants judged that the ML-model predicted correctly in a little bit more than half of the cases (55\%). We found only a slight difference between the perceived usefulness of XAI-methods for the classifications that were judged as correct and those judged as incorrect. The perceived usefulness for the XAI-methods for the headgear study is presented in Table~\ref{tab:usefulness}. In total 115 images were assessed by the participants. The participant could select zero to five methods they found useful. All participants selected at least one XAI-method for each image. Examples of the perceived usefulness are, for example, related to an image picturing a \textquote{stormtrooper helmet} from the Star Wars movies that were classified as a \textquote*{crash helmet}. This was for some users understood as correct since it resembles a \textquote{crash helmet} by the areas accentuated especially by Guided GradCAM. The similarity in functionality was discussed by one user in that a \textquote{stormtrooper helmet} most probable has a protective function but that particular participant also concluded that the superclass \textquote{helmet} would fit better.
In this test, the participant became aware of missing classes, for example, that a \textquote{top hat} was classified as a \textquote*{cowboy hat} (51\% class probability), opened up for these types of conclusions. Other more general opinions were that the ML-model seemed to be good at fabric for example \textquote{wool} but also biased towards water-related headgear (\textquote{bathing cap}, \textquote{bonnet}, \textquote{shower cap}) especially when these were worn by people (spurious correlation).  

\subsection{Concluding remarks on the studies}
It is worth emphasising that the XAI-methods usefulness heavily depends on the human ability to compare the evidence exposed by the XAI-methods with the reality as perceived subjectively by humans. In this setting there is not much to learn about the reality for humans the possible usefulness of the XAI-methods resides in a deepened understanding of the ML-model's inner workings. These insights can then be used to improve the ML-model or delimit its usage domain. It is also worth noting that we only measure perceived usefulness in relation to the other XAI-methods in the study and if all methods were perceived as equally useful, all would end up at 25\% in the animal study and 20\% in the headgear study. The only conclusion we can draw is that, for these mundane images, some methods are perceived as better than others. A deepened study on the mapping between IKR and HUC is needed to evaluate the methods and thus be able to generalise conclusions on their respective usefulness. The lack of a taxonomy that relates concepts to each other also becomes evident in the study, for example classifying an image of a pond with ducks as a \textquote*{drake} without information on that \textquote{drake} is a sub-concept to \textquote{duck}. Here an integration with WordNet~\cite{miller1998wordnet} or a similar service could add further insights for a human.

\section{Discussion}
In this work, we view neural networks as an information processing system that through training builds internal knowledge representations (IKR) inductively based on labels and data in the form of images. As discussed in the background section the trained ML-model is consequently not based on any theory or law in a DN-sense and can therefore not be expected to underpin explanations that hold scientifically. We will in the following discuss what this implies when the intended purpose of XAI-methods is to communicate reasons for a decision in a human understandable fashion. 

We are aware that ImageNet1000 and the ResNet50 model we used are created as a benchmark, primarily focusing on class probability, but we still believe that XAI-methods used on ResNet50 can open up for a discussion on central issues for XAI in general. In the following discussion, we do not include abstract concepts and abstract labels, instead, the focus is on concepts in the form of physical objects clearly visible in the images. The reasoning behind this is that we believe that a baseline for trust in XAI-methods builds on that the methods can, at a minimum, be used to explain promoted decisions for these types of objects. 

It is from our study hard to draw wide-ranging conclusions, except that the two GradCAM methods correlate better with human understanding than the other XAI-methods in the study. The XAI-method Occlusion can be seen as a simpler method since it doesn't focus on IKR directly instead it mechanically measures, somewhat simplified, how important squares in a grid covering image is for a promoted decision. The other methods try, using different approaches, to couple the internal knowledge representations in the ML-model to areas in the image and consequently the proposed decision. As we set up the two GradCAM methods they focus on IKR in the later layers and can thus be useful to understand if high-level sub-concepts (\textquote{ear}, \textquote{horn}) are learned and used when a decision is promoted. The gradient methods that weighs activation, and consequently IKR, over the layers, were in our study deemed as less useful than the other XAI-methods.

To conclude our results, if XAI-methods are used to build trust in the ML-model they need to be used in a structured fashion. Initially, there is a need for a domain expert that can judge the correctness of promoted decisions concerning the intended usage domain in a traditional manner for some test set by using class probability, confusion matrices, etc. Given this starting point, the XAI-methods can be used to deepen the understanding on reasons for decisions. The XAI-methods can be used differently if the promoted decision is perceived as correct or erroneous and consequently the explanation formulated can be associative or contrastive (See the example with \textquote{stormtrooper helmet} in Section \ref{sec:study on headgear}). Another example is the low class probability related to the miss-classification of the \textquote{komodo dragon} as a type of snake \textquote*{Sidewinder} these evidence can together with the focus on the tail by Guided GradCAM be weighed together by a human in order to draw conclusions around the ML-model's abilities.

Making designerly decisions on both the visual appearance and deciding on parameters for the methods, as we did, subjectively influence the results. By focusing on clearly visible mundane objects, a static well-known ML-model and a demographically narrow group of participants we believe that it is possible to draw some conclusions from the study. One is that methods like Occlusion and GradCAM methods are less challenging to understand than the gradient methods. These methods then fit to give an initial oversight over the model's capabilities. 

Below follows a categorisation of the methods in relation to what type of understanding they can communicate in relation to the ML-model's abilities.
\begin{enumerate}
    \item XAI-methods like Occlusion and LIME (builds on initial image segmentation) seems to be the best initial choice to verify that the ML-model focus on the intended areas in the image in relation to the label that is perceived as true. If the ML-model's focus is misaligned with human focus the methods can be used to identify the need for more training data for some classes or to add a label that was previously o.o.d.   
    \item Methods like GradCAM that extracts high-level knowledge representations from the later layers are then useful to identify sub-concepts that should be aligned with human concepts. We believe that these concepts have to be organised and categorised, for example by, using necessary, sufficient concepts and spurious correlations. 
    \item We did not find the different gradient methods useful, more research is then needed to better understand in what way and if they accentuate useful information.   
\end{enumerate}

If the prediction is perceived as correct and the accentuated areas are aligned with the concept important for an explanation, trust will likely increase. For predictions perceived as incorrect areas focused by the XAI-method can be used to draw conclusions either that the object is o.o.d or that the areas accentuated can be used to understand why the image is incorrectly classified. Although this can be important to mitigate bias or find missing classes it is unlikely that by using XAI-methods these ML-models can be used in any ethically challenging setting if the explanandum overlap is not substantial. 
We then believe that XAI-methods can be useful to further inform on reasons for a decision in domains with a substantial explanandum overlap, for example, images picturing objects not reachable for our senses. This can be medical imaging, microscope images, satellite photos, areas where the challenge is hermeneutic and concerns identifying and naming concepts in images. In these cases, the trained ML-model can be seen more as a scientific instrument capable of drawing attention to image details and correlations previously overlooked.

The explanandum overlap in a domain that pictures abstract concepts is limited since the explanandum reachable for the ML-model will be substantially less rich than the explanandum a human has access to. An ML system can still be useful when the goal is to, for example, sort images based on visible concepts like hats, dogs or cats. But as soon as the concepts are abstract and pictures, for example, humans, the system risks being descriptive in coupling physiognomy to concepts. If designers of these systems try to de-bias training data they add their own cultural grounded values and the model becomes normative. Both the descriptive and the normative effects bear risks since the systems can be used globally and the effects on groups that do not share the, by these systems inherited, values are hard to foresee and imagine. These systems then bear similarities with social constructions frozen in time and space and can perhaps be more useful as probes, to compare and understand societal changes in the past compared to the present, than predicting the future.

Even in a limited study like this, the lack of theory and model in a DN-sense related to the knowledge representation in the ML-model becomes evident. For example that three \textquote{mallard ducks} (one drake and two ducks) in a pond were classified as a \textquote*{drake}. The XAI-methods can then draw attention to sub-concepts related to the classification, for example, \textquote{coloured head}, \textquote{tail form}, etc. What evidently is lacking concerns the purpose of the system, why is the \textquote{drake} singled out? To be able to explain a decision there is an obvious need for a contextual grounding and a clear domain delimitation answering why and how questions for the ML system as such. Without clarity concerning this, explanations can only be used to improve the ML-model and not to propose actionable decisions. The capability to name the referent \textquote*{drake} and connect the term to visual concepts severely limits the usability of the systems outside of a biased searching and sorting of existing images. These systems, void of reasoning and understanding, needs at a minimum to connect existing research-grounded taxonomies to their proposed XAI-methods.

The approach we used leaves most of the interpretation of the evidence and, of course, the formulation of explanations to the human using the system. Consequently, selecting explanans to use for an explanation becomes subjective guesswork. For the system to be contestable the XAI-methods needs to be more expressive and not only mark an area as impactful for the decision but also go behind and close in on a why-question related to an impactful sub-concept. We see both the need for local XAI-methods, like those used in this study, and global to be used as an ensemble to understand the strengths and weaknesses of the ML-model. Research in this direction is increasingly attracting attention, by either, forcing the ML-model to learn predefined concepts useful for explanations, or, to extract vectors that bear global information related to the internal knowledge representations~\cite{Akula2020,Koh2020,Mincu2021,Kim2018}. 

Local XAI-methods are then useful when the goal is to falsify an ML-model and global methods when the goal is to get an overview of the intended capabilities concerning the deployment domain. We then need to move towards a situation where we do not mainly optimise on class probability for static datasets and additionally view data drift, bias and inductive learning processes as a solvable technical problem, instead, the main goal must be to view contextual adaptation, continuous learning and a capability to, in a structured fashion, expose human-understandable evidence for a decision. Using an ensemble of XAI-methods can then open up a path to better understand the limitations of the ML-system and thus, perhaps, avoid some of the misuse and over-reliance resulting from a one-sided focus on the ML-systems abilities. 

Finally, we believe that it is time to realise that the emperor is only partly dressed and revisit the scepticism related to pseudoscience one hundred years ago, actualise Hume's problem of induction and apply scientific reasoning on ML/AI~\cite{hempel,inductionproblem}. A clear-eyed view on the state of research in ML/AI is then needed, and, a shift in focus, towards what it cannot do and open up for anthropocentric development with humans in command of a technology that can enhance and/or augment human capabilities.   

\section{Conclusion}
In this paper, we analyse, using our theoretical lens and a case study, consequences of the inductive statistical learning process utilised when a neural networks is trained. Based on this, we identify and discuss the need to complement these systems, if we expect them to be able to create explanations that can be understood. In our study, we especially focus on local XAI-methods and show that reasons for a decision have to be interpreted by a human with domain knowledge in order to be explainable. The XAI-methods used in the study produces vague and incoherent reasons for a decision, reasons that additionally are open for different interpretations. Still, if a domain expert knows the XAI-methods' strengths and weaknesses, they can be useful. 

The usefulness can be increased if XAI-methods can expose reasons for a promoted decision in a human-understandable fashion. Human understandable concepts are then a natural connection point between knowledge internalised in the ML/AI-system and a human aiming to understand a promoted decision. We then strongly believe that benchmarking XAI-methods, using a mundane domain, is a central step in this direction.

Finally, if we are to trust these systems, that profoundly can affect people and societies, we need methods that can be used to challenge promoted decisions. If we cannot develop systems that can explain why an object in an image is an \textquote{elephant} and not a \textquote{zebra} it seems strange to trust them when the stakes are higher. We believe that a shift towards developing ML and AI systems, with an awareness of these systems' epistemological shortcomings, is a point of departure for human in command ML/AI-systems that aims at augmenting and/or enhancing human capabilities.

\begin{acks}
This work was partially financed by the Knowledge Foundation through the Internet of Things and People research profile. I am also in debt to colleagues at my department at Malmö University for their invaluable and insightful comments that substantially improved the work.
\end{acks}
\bibliography{references}

\begin{thebibliography}{44}
\providecommand{\natexlab}[1]{#1}
\providecommand{\url}[1]{\texttt{#1}}
\expandafter\ifx\csname urlstyle\endcsname\relax
  \providecommand{\doi}[1]{doi: #1}\else
  \providecommand{\doi}{doi: \begingroup \urlstyle{rm}\Url}\fi

\bibitem[Adebayo et~al.(2018)Adebayo, Gilmer, Muelly, Goodfellow, Hardt, and
  Kim]{Adebayo2018}
Julius Adebayo, Justin Gilmer, Michael Muelly, Ian Goodfellow, Moritz Hardt,
  and Been Kim.
\newblock {Sanity checks for saliency maps}.
\newblock In \emph{Advances in Neural Information Processing Systems}, volume
  2018-Decem, pages 9505--9515, 2018.
\newblock URL \url{https://goo.gl/hBmhDt}.

\bibitem[Akula et~al.(2020)Akula, Wang, and Zhu]{Akula2020}
Arjun Akula, Shuai Wang, and Song-Chun Zhu.
\newblock {CoCoX: Generating Conceptual and Counterfactual Explanations via
  Fault-Lines}.
\newblock \emph{Proceedings of the AAAI Conference on Artificial Intelligence},
  34\penalty0 (03):\penalty0 2594--2601, 2020.
\newblock ISSN 2159-5399.
\newblock \doi{10.1609/aaai.v34i03.5643}.

\bibitem[Brennan(2017)]{brennan2017}
Andrew Brennan.
\newblock {Necessary and Sufficient Conditions}.
\newblock In Edward~N Zalta, editor, \emph{The Stanford Encyclopedia of
  Philosophy}. Metaphysics Research Lab, Stanford University, summer 2017
  edition, 2017.

\bibitem[Chollet(2019)]{Chollet2019}
François Chollet.
\newblock {On the Measure of Intelligence}.
\newblock \emph{arXiv preprint arXiv:1911.01547}, page~64, 2019.
\newblock URL \url{http://arxiv.org/abs/1911.01547}.

\bibitem[Couldry and Mejias(2019)]{couldry}
Nick Couldry and Ulises~A. Mejias.
\newblock {Data Colonialism: Rethinking Big Data’s Relation to the
  Contemporary Subject}.
\newblock \emph{Television {\&} New Media}, 20\penalty0 (4):\penalty0 336--349,
  2019.
\newblock \doi{10.1177/1527476418796632}.

\bibitem[Deng et~al.(2009)Deng, Dong, Socher, Li, {Kai Li}, and {Li
  Fei-Fei}]{Deng2010}
Jia Deng, Wei Dong, Richard Socher, Li-Jia Li, {Kai Li}, and {Li Fei-Fei}.
\newblock {ImageNet: A large-scale hierarchical image database}.
\newblock In \emph{IEEE conference on computer vision and pattern recognition},
  pages 248--255, Miami, 2009. IEEE.
\newblock \doi{10.1109/cvpr.2009.5206848}.

\bibitem[Gebru et~al.(2021)Gebru, Morgenstern, Vecchione, Vaughan, Wallach,
  Iii, and Crawford]{gebru}
Timnit Gebru, Jamie Morgenstern, Briana Vecchione, Jennifer~Wortman Vaughan,
  Hanna Wallach, Hal~Daumé Iii, and Kate Crawford.
\newblock {Datasheets for datasets}.
\newblock In \emph{Communications of the ACM}, volume~64, pages 86--92, 2021.
\newblock \doi{10.1145/3458723}.

\bibitem[Genone and Lombrozo(2012)]{Genone2012}
James Genone and Tania Lombrozo.
\newblock {Concept possession, experimental semantics, and hybrid theories of
  reference}.
\newblock \emph{Philosophical Psychology}, 25\penalty0 (5):\penalty0 717--742,
  2012.
\newblock ISSN 09515089.
\newblock \doi{10.1080/09515089.2011.627538}.

\bibitem[Gilpin et~al.(2019)Gilpin, Bau, Yuan, Bajwa, Specter, and
  Kagal]{Gilpin2019}
Leilani~H. Gilpin, David Bau, Ben~Z. Yuan, Ayesha Bajwa, Michael Specter, and
  Lalana Kagal.
\newblock {Explaining explanations: An overview of interpretability of machine
  learning}.
\newblock In \emph{Proceedings - 2018 IEEE 5th International Conference on Data
  Science and Advanced Analytics, DSAA 2018}, pages 80--89. IEEE, 2019.
\newblock ISBN 9781538650905.
\newblock \doi{10.1109/DSAA.2018.00018}.

\bibitem[Grimm(2016)]{Grimm2016}
Stephen~R. Grimm.
\newblock {How Understanding People Differs from Understanding the Natural
  World}.
\newblock \emph{Nous-Supplement: Philosophical Issues}, 26\penalty0
  (1):\penalty0 209--225, 2016.
\newblock ISSN 17582237.
\newblock \doi{10.1111/phis.12068}.

\bibitem[Gulrajani and Lopez-Paz(2020)]{gulrajani2020}
Ishaan Gulrajani and David Lopez-Paz.
\newblock {In Search of Lost Domain Generalization}.
\newblock 2020.
\newblock URL \url{http://arxiv.org/abs/2007.01434}.

\bibitem[He et~al.(2016)He, Zhang, Ren, and Sun]{He2016}
Kaiming He, Xiangyu Zhang, Shaoqing Ren, and Jian Sun.
\newblock {Deep residual learning for image recognition}.
\newblock In \emph{Proceedings of the IEEE Computer Society Conference on
  Computer Vision and Pattern Recognition}, volume 2016-Decem, pages 770--778,
  Las Vegas, 2016. IEEE.
\newblock ISBN 9781467388504.
\newblock \doi{10.1109/CVPR.2016.90}.

\bibitem[Hempel and Oppenheim(1948)]{hempel}
Carl~G. Hempel and Paul Oppenheim.
\newblock {Studies in the Logic of Explanation}.
\newblock \emph{Philosophy of Science}, 15\penalty0 (2):\penalty0 135--175,
  1948.
\newblock ISSN 0031-8248.
\newblock \doi{10.1086/286983}.

\bibitem[Henderson(2020)]{inductionproblem}
Leah Henderson.
\newblock {The Problem of Induction}.
\newblock
  https://plato.stanford.edu/archives/spr2020/entries/induction-problem/, 2020.

\bibitem[Holmberg(2021)]{holmberglic}
Lars Holmberg.
\newblock \emph{{Human In Command Machine Learning}}.
\newblock Number~16 in Studies in Computer Science. 2021.
\newblock ISBN 978-91-7877-187-5.
\newblock \doi{10.24834/isbn.9789178771875}.

\bibitem[Hornik(1991)]{Hornik1991}
Kurt Hornik.
\newblock {Approximation capabilities of multilayer feedforward networks}.
\newblock \emph{Neural Networks}, 4\penalty0 (2):\penalty0 251--257, 1991.
\newblock ISSN 08936080.
\newblock \doi{10.1016/0893-6080(91)90009-T}.

\bibitem[Hume(1896)]{Hume1896}
David Hume.
\newblock \emph{{A treatise of human nature}}.
\newblock Clarendon Press, 1896.

\bibitem[Hutchinson and Mitchell(2019)]{Hutchinson2019}
Ben Hutchinson and Margaret Mitchell.
\newblock {50 Years of Test (Un)fairness: Lessons for machine learning}.
\newblock In \emph{FAT* 2019 - Proceedings of the 2019 Conference on Fairness,
  Accountability, and Transparency}, pages 49--58. Association for Computing
  Machinery, Inc, 1 2019.
\newblock ISBN 9781450361255.
\newblock \doi{10.1145/3287560.3287600}.

\bibitem[Kim et~al.(2018)Kim, Wattenberg, Gilmer, Cai, Wexler, Viegas, and
  Sayres]{Kim2018}
Been Kim, Martin Wattenberg, Justin Gilmer, Carrie Cai, James Wexler, Fernanda
  Viegas, and Rory Sayres.
\newblock {Interpretability beyond feature attribution: Quantitative Testing
  with Concept Activation Vectors (TCAV)}.
\newblock In \emph{35th International Conference on Machine Learning, ICML
  2018}, 2018.
\newblock ISBN 9781510867963.

\bibitem[Ko et~al.(2013)Ko, LaToza, and Burnett]{Ko2013a}
Andrew~J. Ko, Thomas~D. LaToza, and Margaret~M. Burnett.
\newblock {A practical guide to controlled experiments of software engineering
  tools with human participants}.
\newblock \emph{Empirical Software Engineering}, 20\penalty0 (1):\penalty0
  110--141, 2013.
\newblock ISSN 15737616.
\newblock \doi{10.1007/s10664-013-9279-3}.

\bibitem[Koh et~al.(2020)Koh, Nguyen, Tang, Mussmann, Pierson, Kim, and
  Liang]{Koh2020}
Pang~Wei Koh, Thao Nguyen, Yew~Siang Tang, Stephen Mussmann, Emma Pierson, Been
  Kim, and Percy Liang.
\newblock {Concept Bottleneck Models}.
\newblock In \emph{International Conference on Machine Learning}, 2020.
\newblock URL \url{https://arxiv.org/abs/2007.04612}.

\bibitem[Kokhlikyan et~al.(2020)Kokhlikyan, Miglani, Martin, Wang, Alsallakh,
  Reynolds, Melnikov, Kliushkina, Araya, Yan, and Reblitz-Richardson]{captum}
Narine Kokhlikyan, Vivek Miglani, Miguel Martin, Edward Wang, Bilal Alsallakh,
  Jonathan Reynolds, Alexander Melnikov, Natalia Kliushkina, Carlos Araya, Siqi
  Yan, and Orion Reblitz-Richardson.
\newblock {Captum: A unified and generic model interpretability library for
  PyTorch}.
\newblock 2020.
\newblock URL \url{http://arxiv.org/abs/2009.07896}.

\bibitem[Lecun et~al.(2015)Lecun, Bengio, and Hinton]{Lecun2015}
Yann Lecun, Yoshua Bengio, and Geoffrey Hinton.
\newblock {Deep learning}.
\newblock \emph{Nature}, 521\penalty0 (7553):\penalty0 436--444, 2015.
\newblock ISSN 14764687.
\newblock \doi{10.1038/nature14539}.

\bibitem[Lipton(2016)]{Lipton2016}
Zachary~C. Lipton.
\newblock {The Mythos of Model Interpretability}.
\newblock \emph{Communications of the ACM}, 61\penalty0 (10):\penalty0 35--43,
  2016.
\newblock ISSN 15577317.
\newblock \doi{10.1145/3233231}.

\bibitem[Lundberg and Lee(2017)]{Lundberg}
Scott~M. Lundberg and Su~In Lee.
\newblock {A unified approach to interpreting model predictions}.
\newblock In \emph{Advances in Neural Information Processing Systems},
  number~3, pages 4766--4775. MIT press, 2017.
\newblock URL \url{https://github.com/slundberg/shap}.

\bibitem[Margolis and Laurence(2021)]{sep-concepts}
Eric Margolis and Stephen Laurence.
\newblock {Concepts}.
\newblock In \emph{The Stanford Encyclopedia of Philosophy}. Metaphysics
  Research Lab, Stanford University, 2021.

\bibitem[Miller(1998)]{miller1998wordnet}
George~A Miller.
\newblock \emph{{WordNet: An electronic lexical database}}.
\newblock MIT press, 1998.

\bibitem[Mincu et~al.(2021)Mincu, Loreaux, Hou, Baur, Protsyuk, Seneviratne,
  Mottram, Tomasev, Karthikesalingam, and Schrouff]{Mincu2021}
Diana Mincu, Eric Loreaux, Shaobo Hou, Sebastien Baur, Ivan Protsyuk, Martin
  Seneviratne, Anne Mottram, Nenad Tomasev, Alan Karthikesalingam, and Jessica
  Schrouff.
\newblock {Concept-based model explanations for electronic health records}.
\newblock In \emph{ACM CHIL 2021 - Proceedings of the 2021 ACM Conference on
  Health, Inference, and Learning}, pages 36--46, 2021.
\newblock \doi{10.1145/3450439.3451858}.

\bibitem[Murphy(2004)]{Murphy2018}
Gregory Murphy.
\newblock \emph{{The Big Book of Concepts}}.
\newblock MIT press, 2004.
\newblock \doi{10.7551/mitpress/1602.001.0001}.

\bibitem[Murphy(2016)]{Murphy2016}
Gregory~L Murphy.
\newblock {Is there an exemplar theory of concepts?}
\newblock \emph{Psychonomic Bulletin and Review}, 23\penalty0 (4):\penalty0
  1035--1042, 2016.
\newblock ISSN 15315320.
\newblock \doi{10.3758/s13423-015-0834-3}.

\bibitem[Myers and Newman(2007)]{Myers2007}
Michael~D. Myers and Michael Newman.
\newblock {The qualitative interview in IS research: Examining the craft}.
\newblock \emph{Information and Organization}, 17\penalty0 (1):\penalty0 2--26,
  2007.
\newblock ISSN 14717727.
\newblock \doi{10.1016/j.infoandorg.2006.11.001}.

\bibitem[Overton(2012)]{Overton}
James~A Overton.
\newblock \emph{{Explanation in Science}}.
\newblock PhD thesis, University of Western Ontario, 2012.
\newblock URL \url{https://ir.lib.uwo.ca/etd/594}.

\bibitem[Samek et~al.(2021)Samek, Montavon, Lapuschkin, Anders, and
  M{\"{u}}ller]{samek2021}
Wojciech Samek, Grégoire Montavon, Sebastian Lapuschkin, Christopher~J.
  Anders, and Klaus~Robert M{\"{u}}ller.
\newblock {Explaining Deep Neural Networks and Beyond: A Review of Methods and
  Applications}.
\newblock \emph{Proceedings of the IEEE}, 109\penalty0 (3):\penalty0 247--278,
  3 2021.
\newblock ISSN 15582256.
\newblock \doi{10.1109/JPROC.2021.3060483}.

\bibitem[Sch{\"{o}}lkopf(2019)]{scholkopf2019}
Bernhard Sch{\"{o}}lkopf.
\newblock {Causality for Machine Learning}.
\newblock 2019.
\newblock URL \url{http://arxiv.org/abs/1911.10500}.

\bibitem[Seawnght and Gerring(2008)]{casestudy}
Jason Seawnght and John Gerring.
\newblock {Case selection techniques in case study research: A menu of
  qualitative and quantitative options}.
\newblock \emph{Political Research Quarterly}, 61\penalty0 (2):\penalty0
  294--308, 2008.
\newblock ISSN 10659129.
\newblock \doi{10.1177/1065912907313077}.

\bibitem[Selvaraju et~al.(2017)Selvaraju, Cogswell, Das, Vedantam, Parikh, and
  Batra]{Selvaraju2016}
Ramprasaath~R. Selvaraju, Michael Cogswell, Abhishek Das, Ramakrishna Vedantam,
  Devi Parikh, and Dhruv Batra.
\newblock {Grad-CAM: Visual Explanations from Deep Networks via Gradient-Based
  Localization}.
\newblock \emph{International Journal of Computer Vision}, 128\penalty0
  (2):\penalty0 336--359, 2017.
\newblock \doi{10.1007/s11263-019-01228-7}.

\bibitem[Sundararajan et~al.(2017)Sundararajan, Taly, and
  Yan]{integratedgradient2017}
Mukund Sundararajan, Ankur Taly, and Qiqi Yan.
\newblock {Axiomatic attribution for deep networks}.
\newblock In \emph{34th International Conference on Machine Learning, ICML
  2017}, volume~7, pages 5109--5118, 2017.
\newblock ISBN 9781510855144.

\bibitem[Tjoa and Guan(2019)]{Tjoa2019}
Erico Tjoa and Cuntai Guan.
\newblock {A Survey on Explainable Artificial Intelligence (XAI): Towards
  Medical XAI}.
\newblock In \emph{IEEE Transactions on Neural Networks and Learning Systems}.
  2019.
\newblock URL \url{https://arxiv.org/abs/1907.07374}.

\bibitem[Vallor(2016)]{Vallor2016}
Shannon Vallor.
\newblock \emph{{Technology and the virtues: A philosophical guide to a future
  worth wanting}}.
\newblock Oxford University Press, 2016.

\bibitem[Wang and Jordan(2021)]{Wang2021}
Yixin Wang and Michael~I Jordan.
\newblock {Desiderata for Representation Learning: A Causal Perspective}.
\newblock 2021.

\bibitem[Wang et~al.(2020)Wang, Mardziel, Datta, and Fredrikson]{Wang2020}
Zifan Wang, Piotr Mardziel, Anupam Datta, and Matt Fredrikson.
\newblock {Interpreting interpretations: Organizing attribution methods by
  criteria}.
\newblock In \emph{IEEE Computer Society Conference on Computer Vision and
  Pattern Recognition Workshops}, volume 2020-June, pages 48--55, 2020.
\newblock ISBN 9781728193601.
\newblock \doi{10.1109/CVPRW50498.2020.00013}.

\bibitem[Woodward and Ross(2021)]{woodward2019}
James Woodward and Lauren Ross.
\newblock {Scientific Explanation}.
\newblock In Edward~N. Zalta, editor, \emph{The Stanford Encyclopedia of
  Philosophy (Summer 2021 Edition)}. 2021.
\newblock URL
  \url{plato.stanford.edu/archives/sum2021/entries/scientific-explanation/}.

\bibitem[Yang et~al.(2020)Yang, Qinami, Fei-Fei, Deng, and
  Russakovsky]{feifei2020}
Kaiyu Yang, Klint Qinami, Li~Fei-Fei, Jia Deng, and Olga Russakovsky.
\newblock {Towards fairer datasets: Filtering and balancing the distribution of
  the people subtree in the ImageNet hierarchy}.
\newblock In \emph{FAT* 2020 - Proceedings of the 2020 Conference on Fairness,
  Accountability, and Transparency}, pages 547--558, 2020.
\newblock ISBN 9781450369367.
\newblock \doi{10.1145/3351095.3375709}.

\bibitem[Zeiler and Fergus(2014)]{occlusion2014}
Matthew~D Zeiler and Rob Fergus.
\newblock {Visualizing and Understanding Convolutional Networks}.
\newblock Technical report, 2014.

\end{thebibliography}
\end{document}